\pdfoutput=1

\documentclass[11pt]{article}

\usepackage[final]{acl}

\usepackage{enumitem}
\usepackage{needspace}
\usepackage{algorithm}
\usepackage{algpseudocode}
\usepackage{amsmath}
\usepackage{booktabs}
\usepackage{multirow}
\usepackage{pifont}
\usepackage{multicol}
\usepackage{caption}
\usepackage{xcolor}
\usepackage{tcolorbox} 
\usepackage{times}
\usepackage{latexsym}
\usepackage{subcaption}
\usepackage{microtype}
\usepackage{tabularx}
\usepackage{inconsolata}
\usepackage{graphicx}
\usepackage{diagbox}
\usepackage{array}

\usepackage[T1]{fontenc}

\usepackage[utf8]{inputenc}

\usepackage{microtype}

\usepackage{inconsolata}

\usepackage{graphicx}

\title{Asking the Right Questions: Ontology-Grounded Interpretable Embeddings for Biomedical Text}

\author{
Yixuan Tang \quad Zhenghong Lin \quad Yandong Sun\thanks{Corresponding author.} \\
{\bf Wynne Hsu \quad Mong Li Lee \quad Anthony Kum Hoe Tung} \\
School of Computing, National University of Singapore \\
\texttt{yixuan@comp.nus.edu.sg \quad zhenghong@u.nus.edu \quad sun.yandong@u.nus.edu} \\
\texttt{\{dcshsuw, dcsleeml, dcstunga\}@nus.edu.sg}
}

\begin{document}
\maketitle


\begin{abstract}

While dense biomedical embeddings achieve strong performance, their opaque dimensions limit transparency in biomedical NLP. Recent question-based interpretable embeddings represent text through binary answers to natural-language questions, but existing approaches rely primarily on corpus-driven signals, often capturing topical or stylistic differences rather than fine-grained biomedical distinctions. We propose \textbf{QIME}, an ontology-grounded framework for interpretable biomedical text embeddings in which each dimension corresponds to an explicit biomedical-domain yes/no question. QIME leverages a biomedical ontology to guide contrastive question generation from semantic clusters, producing atomic, domain-grounded questions. It constructs embeddings via similarity-based semantic activation with MMR-based diversity-aware dimension selection, yielding sparse representations efficiently. Experiments on biomedical clustering, STS and retrieval benchmarks show that QIME consistently outperforms prior interpretable embedding methods and substantially narrows the gap to strong black-box biomedical encoders. It also imposes substantially lower cognitive burden than existing methods, providing concise and domain-specific interpretations. The code is available at \url{https://github.com/L1nzh/QIME}.

\end{abstract}

\section{Introduction}

The deployment of AI systems in high-stakes biomedical applications requires representations that are not only effective but also human-auditable. Specialized dense encoders achieve strong performance across biomedical NLP tasks \cite{DBLP:journals/bioinformatics/LeeYKKKSK20,DBLP:conf/naacl/LiuSMBC21}, but their individual dimensions lack explicit semantic meaning and are therefore difficult to inspect directly \cite{DBLP:conf/emnlp/OpitzMMPC25}. These limitations are especially consequential in downstream health-care settings, where black-box behavior and unreliable post-hoc explanations complicate auditing and failure analysis \cite{Rudin2019Interpretable,Ghassemi2021FalseHope}.

\begin{figure*}[t]
    \centering
    \includegraphics[width=0.95\linewidth]{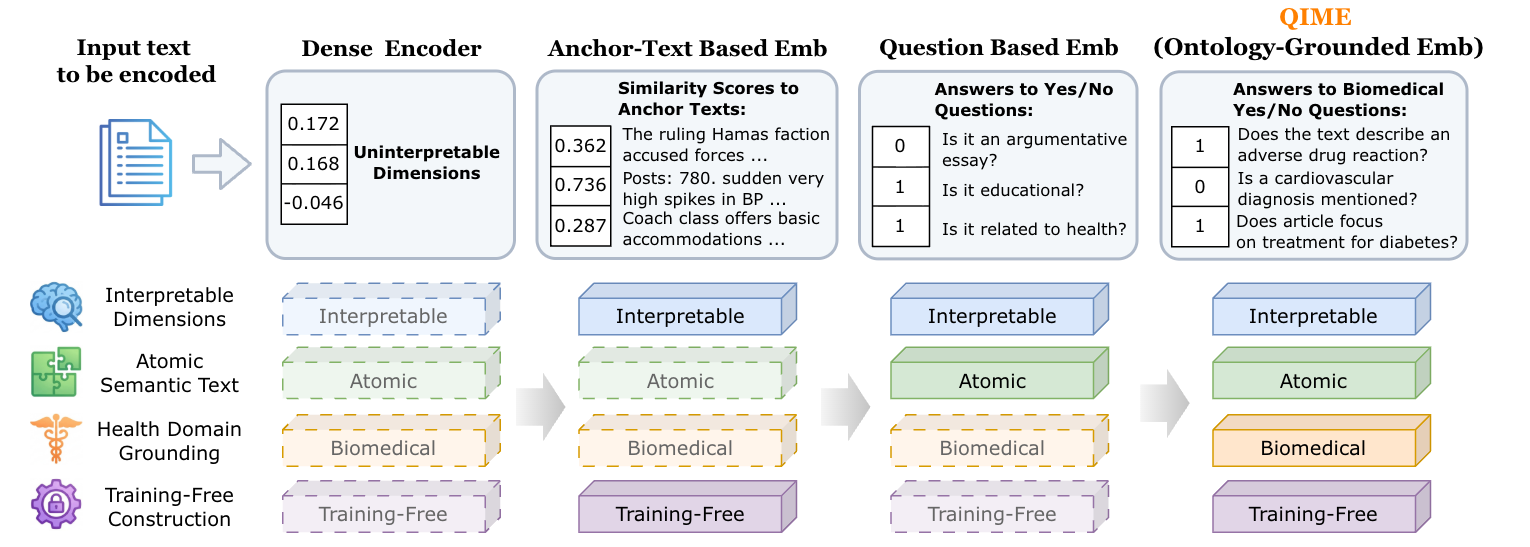}
    \caption{Comparison of text embedding paradigms. Dense encoders produce opaque vectors, anchor-based embeddings rely on heterogeneous reference texts, and question-based embeddings use binary questions derived from corpus-driven signals. \textbf{QIME} grounds question dimensions in biomedical ontology concepts and constructs interpretable embeddings via similarity-based semantic activation with diversity-aware dimension selection.}
    \label{fig:abs}
\end{figure*}


As an illustrative downstream scenario, consider a clinical retrieval system that surfaces similar historical patient cases to support decision-making. With dense embeddings, the system returns only a scalar score with no insight into \textit{why} a case is retrieved. With interpretable embeddings, clinicians can directly inspect the semantic dimensions driving the result. For example, \textit{``Does this case involve CT-based cardiovascular diagnosis?''} or \textit{``Is there evidence of metastatic disease?''} reveal the semantic basis of the retrieval. Such transparency could support auditing and system refinement. When an expected case is missed or an irrelevant case is retrieved, clinicians could identify which activated semantic properties are misaligned. Since QIME's coordinates are explicit questions, task-relevant dimensions can be added to or removed from the question bank, after which document embeddings and the retrieval index can be recomputed without training per-dimension classifiers. This enables targeted refinement that opaque dense coordinates do not directly support, although deployment in clinical workflows would require task-specific validation.

To address this limitation, a growing body of work has explored interpretable text embeddings with human-understandable dimensions \cite{DBLP:conf/emnlp/OpitzMMPC25}. Early efforts include Concept Bottleneck Models (CBMs)~\cite{DBLP:conf/icml/KohNTMPKL20}, which use predefined concepts as intermediate representations. Anchor-based methods~\cite{DBLP:conf/acl/WangSH25} represent texts by similarity to reference documents, but require users to inspect heterogeneous anchors. Recently, question-based embeddings~\cite{DBLP:conf/iclr/SunHTTY25, DBLP:conf/nips/BenaraSMASH024} define each dimension as a binary natural-language question. While this paradigm provides explicit semantic dimensions, generating domain-specific questions from unstructured biomedical corpora remains difficult, often yielding topical or stylistic dimensions rather than fine-grained biomedical distinctions. Existing methods also incur substantial computational overhead due to extensive LLM calls or massive supervised classifiers.

To address these challenges, we introduce \textbf{QIME}, an ontology-grounded framework for constructing question-based interpretable biomedical text embeddings. QIME bridges structured biomedical knowledge and interpretable natural-language representations through ontology-grounded question discovery. Specifically, we cluster a large biomedical corpus, extract biomedical concept signatures for each cluster, and guide an LLM to generate discriminative, domain-specific questions. Next, QIME constructs sparse interpretable embeddings by activating semantically aligned questions as representation coordinates. To reduce redundancy, we further incorporate diversity-aware dimension selection via Maximal Marginal Relevance (MMR). This avoids training massive per-question classifiers and enables scalable construction of interpretable embeddings. Figure~\ref{fig:abs} shows the distinctions among embedding paradigms.

We evaluate QIME on biomedical benchmarks spanning semantic textual similarity, clustering, and information retrieval. Results show that QIME outperforms prior interpretable embedding methods and substantially narrows the gap to strong black-box biomedical encoders. Qualitative analyses further demonstrate that QIME provides explicit, ontology-grounded dimensions.

Our main contributions are:
\begin{itemize}[itemsep=0.5pt, topsep=0.5pt]
    \item We propose \textbf{QIME}, an \textbf{ontology-grounded} framework for constructing \textbf{question-based interpretable biomedical text embeddings} with explicit, domain-specific dimensions. 
    \item We introduce a \textbf{training-free similarity-based semantic activation} mechanism for constructing sparse interpretable embeddings, together with \textbf{MMR-based diversity-aware dimension selection} to reduce redundancy among activated questions.
    \item We show that QIME achieves \textbf{strong empirical performance} on the biomedical STS task in MTEB and biomedical clustering and retrieval benchmarks, and \textbf{reduces cognitive load} by over 76\% compared to existing question-based encoders. 
    
\end{itemize}

\section{Related Work}
\label{sec:related}

\begin{figure*}[t]
    \centering
    \includegraphics[width=0.95\linewidth]{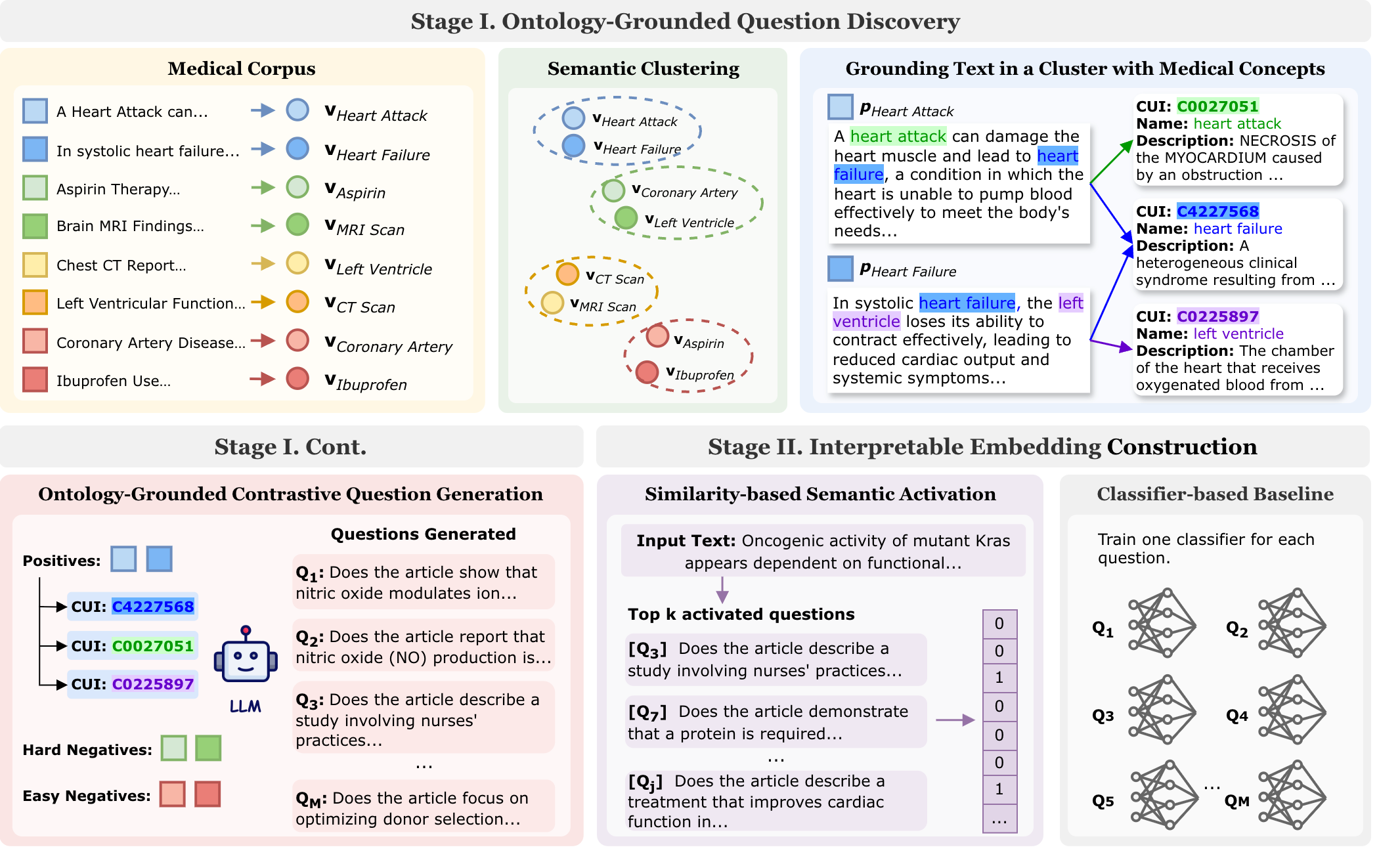}
    \caption{Overview of the QIME framework.}
    \label{fig:overview}
\end{figure*}

\subsection{Black-Box Text Embeddings}

Dense neural embeddings dominate modern NLP pipelines for semantic similarity, clustering, and retrieval. Contextual encoders such as BERT \cite{DBLP:conf/naacl/DevlinCLT19} and contrastive sentence models like SimCSE \citep{DBLP:conf/emnlp/GaoYC21} achieve strong performance. Recent work shows that decoder-only language models can also be repurposed as embedding models \cite{llm2vec,DBLP:journals/corr/abs-2509-20354}.

In the biomedical domain, continued pretraining yields specialized encoders, including BioBERT \cite{DBLP:journals/bioinformatics/LeeYKKKSK20}, and PubMedBERT \citep{DBLP:journals/health/GuTCLULNGP22}. Ontology-aware models, SapBERT \citep{DBLP:conf/naacl/LiuSMBC21} and BioLORD \citep{DBLP:journals/jamia/RemyDD24}, leverage the Unified Medical Language System (UMLS) \cite{DBLP:journals/nar/Bodenreider04} synonym sets to improve biomedical representations. Despite their effectiveness, these models produce dense representations whose dimensions lack explicit semantic meaning. 

\subsection{Interpretable Text Embeddings}

To address the opacity of dense encoders, prior work has explored interpretable representations \cite{DBLP:conf/emnlp/OpitzMMPC25}. Concept-based approaches, such as CBMs \cite{DBLP:conf/icml/KohNTMPKL20}, TCAV \cite{DBLP:conf/icml/KimWGCWVS18}, and BIERs \cite{DBLP:conf/acl/Garcia-OlanoOBG21}, rely on predefined or weakly supervised concepts and offer limited flexibility when new concepts are added or modified. 

A more recent direction is question-based embeddings, where dimensions are associated with binary natural-language questions. QA-Emb \cite{DBLP:conf/nips/BenaraSMASH024} uses LLM prompting to generate interpretable features, but requires querying the LLM for all dimensions at inference time. CQG-MBQA \cite{DBLP:conf/iclr/SunHTTY25} introduces Contrastive Question Generation to produce discriminative questions from semantic clusters, but relies solely on corpus-level distributional signals, yielding generic questions that lack biomedical-domain specificity. Moreover, it relies on training a dedicated classifier for each dimension, introducing additional annotation and training overhead. Anchor-based methods such as LDIR \cite{DBLP:conf/acl/WangSH25} represent texts via similarity to anchors, achieving compact representations but requiring users to interpret dimensions through long, heterogeneous anchor texts rather than self-describing semantic units.

In contrast, QIME grounds question generation in UMLS concept signatures, constructs embeddings via training-free semantic activation, and applies diversity-aware dimension selection via MMR to yield sparse, complementary, domain-specific, and interpretable representations.

\section{The QIME Framework} 
\label{sec:method}

\subsection{Overview}

We propose \textbf{QIME} (Ontology-Grounded \textbf{Q}uestion-based \textbf{I}nterpretable bio\textbf{M}edical \textbf{E}mbeddings), a framework for constructing interpretable embeddings for biomedical text in which each dimension corresponds to an explicit biomedical-domain yes/no question (e.g., \emph{``Does the text describe adverse drug reactions?''}). Formally, given a biomedical text corpus $\mathcal{D} = \{x_i\}_{i=1}^N$, QIME aims to learn an embedding function $f : x \mapsto \mathbf{z} \in \{0,1\}^M$, where $z_j \in \{0,1\}$. The binary value indicates whether a question dimension is semantically activated for the document. Unlike prior question-based approaches, questions are not predefined or heuristic but automatically discovered through an ontology-grounded contrastive prompting process.

As illustrated in Figure~\ref{fig:overview}, the QIME framework consists of two key stages:
(1) \textbf{Ontology-grounded question generation}, which discovers explicit biomedical-domain question dimensions from an unlabeled biomedical corpus;
(2) \textbf{Interpretable embedding construction}, which encodes new texts into sparse question-indexed representations via similarity-based semantic activation and diversity-aware dimension selection through MMR, without requiring QA supervision or classifier training.

\subsection{Ontology-Grounded Question Discovery}

The objective of the first stage is to discover a set of questions that are both discriminative with respect to the corpus and grounded in biomedical ontology concepts. Purely data-driven question generation often produces surface-level or stylistic distinctions, which are inadequate for domain-specific interpretation. QIME addresses this issue by combining corpus-level semantic structure with explicit ontology grounding.

\paragraph{Semantic Clustering of the Biomedical Corpus.}

We first cluster the corpus into semantically coherent groups. Each document $x_i$ is encoded into a dense representation $\mathbf{h}_i$. The corpus is then clustered into $K$ clusters, $\mathcal{D} = \bigcup_{k=1}^K \mathcal{C}_k$. Each cluster typically corresponds to a latent biomedical topic or concept region, such as diagnoses, treatments, or medications. 

\paragraph{Cluster-Level Ontology Grounding.}
To align the discovered semantic clusters with established domain knowledge, we ground each cluster in a biomedical ontology. For a given cluster $\mathcal{C}_k$, we apply named entity recognition and entity linking to all documents in the cluster to identify biomedical entities, which are then mapped to ontology concepts. Specifically, we use Concept Unique Identifiers (CUIs) from the 2025AA release of the Unified Medical Language System (UMLS) \cite{DBLP:journals/nar/Bodenreider04,nlm2025umls}, a large-scale biomedical ontology, where each CUI represents a canonical biomedical concept that unifies synonymous terms across different biomedical vocabularies. The CUIs extracted from cluster $\mathcal{C}_k$ are aggregated to form a cluster-level concept signature $\mathcal{U}_k = \{u_1, u_2, \dots, u_{|\mathcal{U}_k|}\}$.

\paragraph{Ontology-Grounded Question Discovery.}

Given a target cluster $\mathcal{C}_k$ and its concept signature $\mathcal{U}_k$, QIME generates binary biomedical-domain questions capturing the cluster's defining semantic properties via a contrastive prompting strategy. For each cluster, we construct three types of examples: \textbf{positive samples} drawn from $\mathcal{C}_k$; \textbf{hard negatives} from semantically nearest clusters (by centroid distance); and \textbf{easy negatives} from randomly selected distant clusters, sampling six examples from each selected cluster.





An LLM is prompted to generate yes/no questions that distinguish positives from both negative types, explicitly conditioned on the ontology concepts in $\mathcal{U}_k$, including concept names and descriptions.
By jointly leveraging contrastive supervision and ontology constraints, the generated questions are encouraged to reflect domain-specific biomedical distinctions. Generated questions are aggregated and post-processed to remove low-quality, ambiguous, and redundant entries, resulting in a final set of $M$ questions, $\mathcal{Q} = \{q_1, \dots, q_M\}$. Prompts and post-processing details are in Appendices \ref{sec:appendix_prompt} and~\ref{sec:appendix_post}.

\subsection{Interpretable Biomedical Embedding Construction}

Once the question set $\mathcal{Q}$ is obtained, the second stage constructs interpretable embeddings for individual documents. 

\begin{table*}[t]
\centering
\small
\begin{tabular}{ccccccccc}
\toprule
\multirow{2}{*}{\textbf{Type}} & 
\multirow{2}{*}{\textbf{Model}} & 
\multicolumn{6}{c}{\textbf{Clustering ( V-Measure $\uparrow$ )}} & \textbf{STS ( SC $\uparrow$ )} \\
\cmidrule(lr){3-8} \cmidrule(lr){9-9}
& &
\textbf{BioP2P} &
\textbf{BioS2S} &
\textbf{MedP2P} &
\textbf{MedS2S} &
\textbf{ClusTREC} &
\textbf{Average} &
\textbf{BIOSSES} \\
\midrule
\multirow{11}{*}{\rotatebox{90}{\textbf{Black-Box}}} &

\textbf{BERT} & 29.95 & 24.40 & 26.13 & 23.63 & 74.50 & 35.72 & 54.70 \\
& \textbf{GloVe} & 29.32 & 18.74 & 26.14 & 20.49 & 74.15 & 33.77 & 44.93 \\
& \textbf{SimCSE (Unsup)} & 30.10 & 22.94 & 28.03 & 25.62 & 76.41 & 36.62 & 68.86 \\
& \textbf{SimCSE (Sup)} & 31.91 & 25.70 & 28.38 & 25.85 & 76.54 & 37.68 & 67.19 \\

& \textbf{MedEmbed} & \textbf{40.10} & \textbf{35.99} & \textbf{33.12} & \underline{30.44} & \textbf{83.26} & \textbf{44.58} & \underline{86.99} \\
& \textbf{EmbeddingGemma} & \underline{36.95} & \underline{33.06} & 31.68 & \textbf{30.45} & 82.57 & \underline{42.94} & 80.46 \\
& \textbf{PubMedBERT} & 34.37 & 30.97 & \underline{32.36} & 28.12 & \underline{82.59} & 41.68 & 83.96 \\
& \textbf{BioLORD} & 31.30 & 27.87 & 31.77 & 30.28 & 80.03 & 40.25 & \textbf{87.18} \\
& \textbf{SapBERT} & 31.00 & 20.53 & 29.43 & 22.86 & 77.05 & 36.17 & 82.48 \\
& \textbf{MedCPT} & 35.11 & 32.74 & 30.49 & 29.29 & 77.77 & 41.08 & 81.95 \\
& \textbf{BMRetriever} & 34.48 & 20.34 & 29.81 & 22.62 & 79.39 & 37.33 & 68.85 \\

\midrule
\multirow{5}{*}{\rotatebox{90}{\shortstack{\textbf{Inter-}\\\textbf{pretable}}}} &
\textbf{Bag-of-Words} & 4.73 & 3.32 & 12.43 & 13.05 & 65.68 & 19.84 & 68.78 \\
& \textbf{LDIR-500} & 32.39 & 29.36 & 30.00 & \underline{28.98} & 79.54 & 40.05 & \underline{79.30} \\
& \textbf{CQG-MBQA} & \underline{34.88} & \underline{31.14} & \underline{31.02} & 28.65 & \underline{79.67} & \underline{41.07} & 54.97 \\
& \textbf{QA-Emb} & 24.60 & 21.11 & 25.53 & 22.82 & 75.30 & 33.87 & 46.43 \\
& \textbf{QIME} & \textbf{40.37} & \textbf{36.78} & \textbf{33.92} & \textbf{31.44} & \textbf{81.99} & \textbf{44.90} & \textbf{79.66} \\

\bottomrule
\end{tabular}

\caption{Clustering (V-Measure) and semantic textual similarity (Spearman correlation) results on biomedical benchmarks. Best results are in bold and second-best results are underlined within each category.}

\label{tab:main_clus}
\end{table*}

\paragraph{Similarity-Based Semantic Activation.}

QIME constructs embeddings through a similarity-based semantic activation mechanism that measures the semantic alignment between an input document and each question dimension. Each question acts as an interpretable semantic probe, and dimensions are activated based on their relevance to the document, producing sparse and interpretable representations without requiring explicit QA supervision. Given a document $x$, we encode it into a dense vector $\mathbf{h}(x)$ and encode all questions into $\{\mathbf{h}(q_j)\}_{j=1}^M$. We compute cosine similarity scores $s_j = \mathrm{sim}(\mathbf{h}(x), \mathbf{h}(q_j))$ and activate the top-$k$ dimensions, where $k \ll M$:
\[
z_j =
\begin{cases}
1, & \text{if } q_j \in \mathrm{Top}\text{-}k(s_1, \dots, s_M), \\
0, & \text{otherwise}.
\end{cases}
\]

To reduce redundancy among highly similar questions, QIME further incorporates Maximal Marginal Relevance (MMR) during the selection process. Let $s_{i,x}=\mathrm{sim}(\mathbf{h}(q_i),\mathbf{h}(x))$ denote document relevance and $s_{i,j}=\mathrm{sim}(\mathbf{h}(q_i),\mathbf{h}(q_j))$ denote inter-question similarity. We select questions using
\[
\begin{split}
q^* = \arg\max_{q_i \in \mathcal{Q} \setminus S} [&
\lambda s_{i,x} - (1-\lambda) \max_{q_j \in S} s_{i,j}].
\end{split}
\]
Here, $S$ is the set of already-selected questions. Questions are chosen iteratively by jointly maximizing relevance to the document and dissimilarity to previously selected questions, encouraging activated dimensions to capture complementary semantic aspects of the input. By activating a small set of diverse questions, QIME produces concise and interpretable embeddings.

\paragraph{Computational Efficiency.} Compared to prior question-based methods, QIME's embedding construction is substantially more efficient. QA-Emb requires $M$ separate LLM queries at inference time, while classifier-based methods such as CQG-MBQA require training $M$ dedicated classifiers (requiring millions of annotated instances) and querying all $M$ at inference. QIME requires no training, and inference consists of a single encoder forward pass, one matrix multiplication between the document and all $M$ question embeddings, and $O(k \times M)$ lightweight cosine computations for MMR selection.

\section{Experiments}
\label{sec:experiments}

\begin{table*}[th]
\centering
\small
\begin{tabular}{ccccccccc}
\toprule
\multirow{2}{*}{\textbf{Type}} &
\multirow{2}{*}{\textbf{Model}} &
\multicolumn{7}{c}{\textbf{Retrieval ( nDCG@10 $\uparrow$ )}} \\
\cmidrule(lr){3-9}
& &
\textbf{NFCorpus} &
\textbf{PHQA} &
\textbf{MedQA} &
\textbf{COVID} &
\textbf{R2-IYI} &
\textbf{R2-PMC} &
\textbf{Average} \\
\midrule

\multirow{11}{*}{\rotatebox{90}{\textbf{Black-Box}}} &
\textbf{BERT} & 4.30 & 46.20 & 9.78 & 14.78 & 6.90 & 1.80 & 13.96 \\
& \textbf{GloVe} & 13.87 & 62.57 & 19.95 & 36.22 & 7.88 & 7.31 & 24.63 \\
& \textbf{SimCSE (Unsup)} & 9.88 & 61.07 & 24.51 & 32.71 & 10.07 & 6.43 & 24.11 \\
& \textbf{SimCSE (Sup)} & 12.42 & 65.89 & 24.27 & 30.83 & 8.28 & 4.94 & 24.44 \\
& \textbf{EmbeddingGemma} & \underline{31.42} & \underline{78.70} & 60.05 & 50.36 & 12.66 & 9.20 & \underline{40.40} \\
& \textbf{MedEmbed} & \textbf{37.07} & \textbf{82.37} & \textbf{74.82} & \textbf{75.73} & \textbf{14.96} & \underline{11.25} & \textbf{49.37} \\
& \textbf{PubMedBERT} & 26.60 & 68.42 & 58.01 & 44.76 & \underline{12.77} & \textbf{12.51} & 37.18 \\
& \textbf{BioLORD} & 25.49 & 74.77 & \underline{61.49} & \underline{54.89} & 12.22 & 6.09 & 39.16 \\
& \textbf{SapBERT} & 26.77 & 57.38 & 58.45 & 33.40 & 9.27 & 5.48 & 31.79 \\
& \textbf{MedCPT} & 28.43 & 53.97 & 40.46 & 54.66 & 6.09 & 8.04 & 31.94 \\
& \textbf{BMRetriever} & 3.04 & 38.62 & 10.15 & 18.64 & 11.22 & 10.79 & 15.41 \\

\midrule
\multirow{5}{*}{\rotatebox{90}{\shortstack{\textbf{Inter-}\\\textbf{pretable}}}} & 
\textbf{Bag-of-Words} & 21.59 & 42.29 & 26.01 & 19.23 & 6.83 & 4.86 & 20.14 \\
& \textbf{LDIR-500} & \textbf{27.08} & \underline{70.68} & \textbf{65.69} & \underline{47.04} & \textbf{13.39} & \textbf{10.87} & \underline{39.13} \\
& \textbf{CQG-MBQA} & 9.74 & 62.27 & 40.45 & 28.49 & 10.58 & 6.88 & 26.40 \\
& \textbf{QA-Emb} & 3.87 & 44.95 & 21.71 & 22.42 & 9.10 & 5.11 & 17.86 \\
& \textbf{QIME} & \underline{25.09} & \textbf{75.64} & \underline{62.36} & \textbf{64.65} & \underline{11.79} & \underline{7.08} & \textbf{41.10} \\

\bottomrule
\end{tabular}
\caption{Retrieval results (nDCG@10) across biomedical benchmarks.}

\label{tab:main_retrieval}
\end{table*}

\subsection{Experimental Setup}
\label{sec:setup}

\paragraph{Tasks and Datasets.}
We evaluate interpretable embeddings on three embedding-centric biomedical NLP tasks: (i) text clustering, (ii) semantic textual similarity (STS), and (iii) text retrieval, following the widely adopted text embedding evaluation protocol established by MTEB \citep{DBLP:conf/eacl/MuennighoffTMR23}. The tasks jointly assess the full spectrum of embedding quality: topical structure discovery, fine-grained semantic alignment, and query-document matching.

For \textbf{clustering}, we use biomedical subsets from the Massive Text Embedding Benchmark (MTEB) \citep{DBLP:conf/eacl/MuennighoffTMR23}, including BiorxivClusteringP2P (BioP2P), BiorxivClusteringS2S (BioS2S), MedrxivClusteringP2P (MedP2P), and MedrxivClusteringS2S (MedS2S).
They require grouping biomedical preprints based on either titles or abstracts. We additionally evaluate ClusTREC-Covid (ClusTREC) \citep{DBLP:conf/emnlp/KatzLG24}, a COVID-focused clustering benchmark derived from TREC-COVID literature. We report V-Measure for all clustering tasks.

For \textbf{STS}, we use BIOSSES \citep{DBLP:conf/eacl/MuennighoffTMR23}, which contains 100 biomedical sentence pairs annotated for semantic relatedness on a 0--4 scale. We report Spearman correlation for STS.

For \textbf{retrieval}, we evaluate NFCorpus \citep{DBLP:conf/ecir/BotevaGSR16} and TREC-COVID (COVID) \citep{DBLP:journals/sigir/VoorheesABDHLRS20}. We further include medical QA retrieval benchmarks PublicHealthQA (PHQA) \citep{DBLP:conf/iclr/EnevoldsenCKKMS25} and MedicalQARetrieval (MedQA) \citep{DBLP:journals/bmcbi/AbachaD19}, as well as reasoning-intensive clinical retrieval from R2MED \citep{DBLP:journals/corr/abs-2505-14558}, using its MTEB variants R2MEDIIYiClinicalRetrieval (R2-IYI) and R2MEDPMCClinicalRetrieval (R2-PMC). nDCG@10 is reported for all datasets.

\begin{table*}[t]
\centering
\small
\begin{tabular}{lccccccc}
\toprule
\multirow{2}{*}{\textbf{Model}} & \multicolumn{6}{c}{\textbf{Clustering (V-Measure $\uparrow$)}} & \textbf{STS (SC $\uparrow$)} \\
\cmidrule(lr){2-7} \cmidrule(lr){8-8}
& \textbf{BioP2P} & \textbf{BioS2S} & \textbf{MedP2P} & \textbf{MedS2S} & \textbf{ClusTREC} & \textbf{Average} & \textbf{BIOSSES} \\
\midrule

\textbf{QIME}  & 40.37 & 36.78 & 33.92 & 31.44 & 81.99 & 44.90 & 79.66 \\
\textbf{QIME w/o MMR} & 40.26 & 36.83 & 33.78 & 31.83 & 81.69 & 44.88 & 75.60 \\
\textbf{QIME w/o ontology}  & 38.75 & 35.60 & 33.36 & 31.21 & 81.21 & 44.03 & 74.32 \\
\textbf{QIME\_CLS}  & 38.18 & 34.82 & 33.61 & 32.00 & 79.43 & 43.61 & 61.88 \\
\textbf{QIME\_CLS w/o ontology}
 & 37.29 & 34.01 & 32.91 & 31.12 & 79.22 & 42.91 & 57.77 \\

\midrule

& \multicolumn{7}{c}{\textbf{Retrieval (nDCG@10 $\uparrow$)}} \\
\cmidrule(lr){2-8}
& \textbf{NFCorpus} & \textbf{PHQA} & \textbf{MedQA} & \textbf{COVID} & \textbf{R2-IYI} & \textbf{R2-PMC} & \textbf{Average} \\
\midrule

\textbf{QIME}  & 25.09 & 75.64 & 62.36 & 64.65 & 11.79 & 7.08 & 41.10 \\
\textbf{QIME w/o MMR} & 21.29 & 75.04 & 57.79 & 57.96 & 10.09 & 5.99 & 38.03 \\
\textbf{QIME w/o ontology}
 & 20.76 & 74.13 & 58.70 & 54.29 & 11.00 & 7.27 & 37.69 \\
\textbf{QIME\_CLS}  & 15.74 & 61.74 & 54.66 & 46.31 & 8.94 & 5.53 & 32.15 \\
\textbf{QIME\_CLS w/o ontology}
& 15.35 & 61.07 & 51.45 & 44.56 & 8.82 & 4.39 & 30.94 \\

\bottomrule
\end{tabular}
\caption{Ablation study analyzing the impact of each component of QIME.} 
\label{tab:ablation}
\end{table*}

\begin{figure*}[th]
    \centering
    \includegraphics[width=\linewidth]{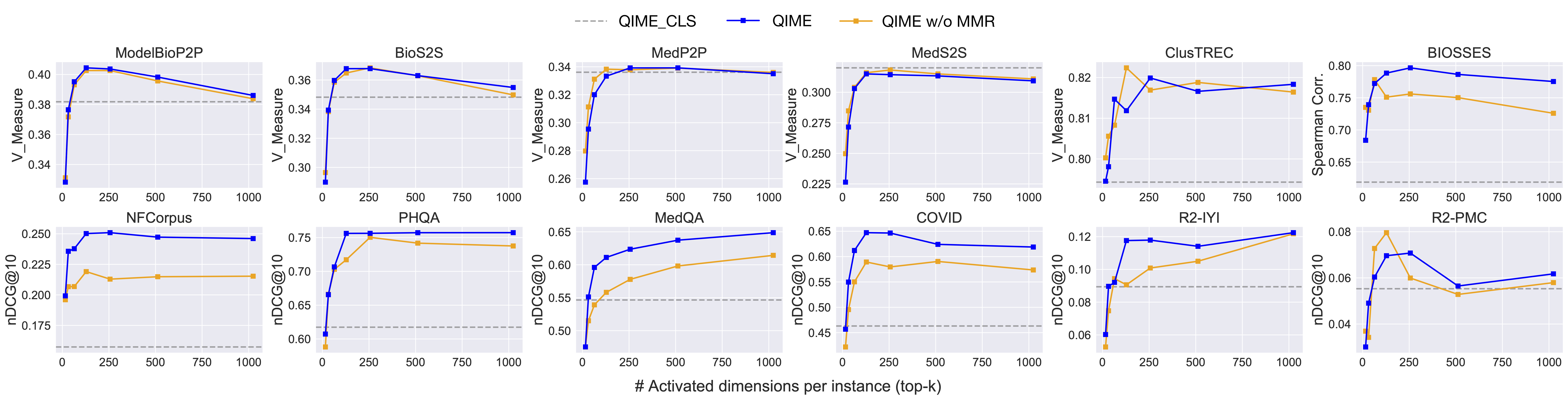}
    \caption{Effect of the top-$k$ dimension selection in similarity-based semantic activation. Performance peaks at moderate sparsity (k=128–256), and MMR-based selection improves results.}
    \label{fig:topk}
\end{figure*}

\paragraph{Baselines.}
We compare QIME against strong black-box and interpretable baselines.

For \textbf{black-box dense encoders}, we include general-domain models BERT \cite{DBLP:conf/naacl/DevlinCLT19}, GloVe \citep{DBLP:conf/emnlp/PenningtonSM14}, supervised and unsupervised SimCSE \citep{DBLP:conf/emnlp/GaoYC21}, and the decoder-based embedding model EmbeddingGemma \citep{DBLP:journals/corr/abs-2509-20354}. To assess domain-specific performance, we evaluate biomedical encoders PubMedBERT \citep{DBLP:journals/health/GuTCLULNGP22}, SapBERT \citep{DBLP:conf/naacl/LiuSMBC21}, and BioLORD-2023 \citep{DBLP:journals/jamia/RemyDD24}, as well as retrieval-oriented models MedCPT \citep{DBLP:journals/bioinformatics/JinKCCYWL23}, BMRetriever \citep{DBLP:conf/emnlp/0002SYZ0WHZ024}, and MedEmbed \citep{balachandran2024medembed}.

For \textbf{interpretable embeddings}, we include a bag-of-words baseline \citep{DBLP:journals/ipm/SaltonB88}, question-based embeddings QA-Emb \citep{DBLP:conf/nips/BenaraSMASH024} and CQG-MBQA \citep{DBLP:conf/iclr/SunHTTY25}, as well as LDIR-500 \citep{DBLP:conf/acl/WangSH25}, which represents texts via relative similarities to a fixed set of $500$ diverse anchor texts. The main tables compare complete published systems under their own reported best configurations.

\paragraph{Implementation Details.}

We preprocess the widely adopted PubMed corpus \cite{roberts2001pubmed} released with MedRAG \citep{xiong2024benchmarking}, randomly sampling 5M out of 25M paragraphs (average length 296 tokens) for semantic clustering after filtering low-quality and duplicated entries. Documents are encoded using MedEmbed \citep{balachandran2024medembed} and clustered by k-means ($K=2{,}500$). Biomedical entity mentions are extracted using HunFlair2 \citep{saenger2024hunflair2} and mapped to the UMLS 2025AA release \citep{nlm2025umls}.

For the cluster-level concept signature, we retain the five most frequent mapped concepts for each cluster as $\mathcal{U}_k$ to ensure focus. Qwen3-30B \citep{DBLP:journals/corr/abs-2505-09388} is used as the LLM backbone for grounded question generation. After post-processing, 8,855 questions are retained. We set $\lambda = 0.7$ in MMR and $k = 256$ in the top-$k$ dimension activation.



\subsection{Main Results: Embedding Quality}
\label{sec:main-results}

\paragraph{Clustering and STS.}

Table~\ref{tab:main_clus} reports results on clustering and STS benchmarks. Strong black-box models such as MedEmbed, EmbeddingGemma, and BioLORD achieve the strongest performance. Notably, QIME surpasses all black-box encoders on 4 out of 5 clustering tasks, a strong outcome for an interpretable method.

Within the interpretable category, QIME consistently outperforms interpretable baselines across all five clustering tasks and on BIOSSES. In particular, QIME substantially improves over QA-Emb and CQG-MBQA, indicating that ontology-grounded question discovery produces more coherent and discriminative semantic dimensions than manually designed or purely data-driven questions.

\paragraph{Retrieval.}
Table~\ref{tab:main_retrieval} reports retrieval performance (nDCG@10). Among interpretable methods, QIME achieves the highest average score, outperforming CQG-MBQA on every dataset and obtaining a higher average than LDIR-500 (41.10 vs.\ 39.13), although their relative performance varies across datasets. The comparison with LDIR-500 reflects different representation trade-offs: its 500 continuous anchor similarities may preserve fine-grained ranking signals, whereas QIME activates at most 256 binary question dimensions per document, favoring sparse and directly inspectable representations. This qualitative difference is illustrated in Table~\ref{tab:case_study}.

\paragraph{Overall.}
Across all three task types, QIME outperforms prior interpretable methods and substantially narrows the gap to dense embeddings, without per-question classifier training or test-time LLM inference. Additional experiments with alternative LLM backbones (LLaMA-3-70B and GPT-5-mini) show similar trends, indicating that QIME is robust to the choice of backbone (Appendix~\ref{sec:appendix_backbone}).

\begin{table*}[t]
\centering
\small
\begin{tabular}{cccccccc}
\toprule
\textbf{Model} & \textbf{NFCorpus} & \textbf{PHQA} & \textbf{MedQA} & \textbf{COVID} & \textbf{R2-IYI} & \textbf{R2-PMC} & \textbf{Average} \\

\midrule
\textbf{QA-Emb}   & 1,515 & 1,883 & 1,682 & 2,071 & 2,191 & 2,106 & 1,908 \\
\textbf{CQG-MBQA}  & 479  & 512  & 529  & 543  & 722  & 713  & 583  \\
\textbf{QIME}      & 101  & 115  & 121  & 150  & 180  & 152  & 136  \\
\bottomrule
\end{tabular}
\caption{Cognitive load (lower is better) across retrieval datasets. QIME imposes substantially less cognitive burden than existing question-based interpretable encoders.}
\label{tab:cognitive_load}
\end{table*}

\begin{table*}[t]
\centering
\small
\begin{tabular}{ccp{0.75\linewidth}}
\multicolumn{3}{p{0.95\linewidth}}{\textbf{Input:}
\textit{The patient, with a history of lung cancer, presented with chest pain. Initial concern for heart attack was raised, but a contrast-enhanced CT scan revealed no coronary occlusion. It showed metastatic disease involving the mediastinum.}} \\
\toprule
\textbf{Model} & \textbf{Rank} & \textbf{Content of the top-rated Dimensions} \\
\midrule
\multirow{3}{*}{\textbf{LDIR-500}}
& 1 & I had to get a chest 
recently for severe chest pains. My period was due one week from ... Two weeks later, I was a week late with my period and I had a positive pregnancy test. \\
& 2 & Thickening of the pleural membranes is not a condition which is treatable. It is a symptom of a disease such as asbestosis, treatment is more focused on the underlying cause of the thickening. \\
& 3 & Cancer June 21 -- July 22. People bearing the Cancer sign are so loving, you can almost consider them emotional. Cancers make up the greater part of caring folks on this earth. \\
\midrule

\multirow{3}{*}{\textbf{CQG-MBQA}}
& 1 & Does the article discuss medical conditions or treatments? \\
& 2 & Is the article devoid of promoting extreme diets or quick fixes? \\
& 3 & Does the article provide factual information rather than anecdotal stories? \\
\midrule

\multirow{3}{*}{\textbf{QIME}}
& 1 & Is there a focus on pain control in terminally ill or cancer patients? \\
& 2 & Does the article involve the use of computed tomography (CT) for diagnosing cardiovascular conditions in humans? \\
& 3 & Is there evidence of a bacterial infection affecting heart valves or cardiac tissue? \\
\bottomrule
\end{tabular}

\caption{Qualitative comparison of top-ranked interpretable dimensions across methods for the same clinical input. LDIR returns lengthy, noisy anchor texts, while CQG-MBQA produces generic questions. QIME activates concise, ontology-grounded questions that are easier to inspect and more closely aligned with the clinical themes of this input.}



\label{tab:case_study}
\end{table*}

\subsection{Ablation Study}
\label{sec:ablation}

Table~\ref{tab:ablation} analyzes three components of QIME: embedding construction strategy, ontology grounding for question generation, and diversity-aware dimension selection via MMR. To isolate these factors, we evaluate four variants: (i) \textbf{QIME w/o MMR}, which removes diversity-aware selection; (ii) \textbf{QIME w/o ontology}, which disables ontology grounding during question generation; (iii) \textbf{QIME\_CLS}, a classifier-based embedding construction strategy where each dimension is predicted by a dedicated binary classifier (details in Appendix~\ref{sec:appendix_cls}); and (iv) \textbf{QIME\_CLS w/o ontology}, which further removes ontology grounding in QIME\_CLS.

\paragraph{Effect of Embedding Construction Strategy.}
We first compare the proposed similarity-based semantic activation with the classifier-based variant QIME\_CLS. Across clustering, STS, and retrieval tasks, QIME consistently achieves better average results than QIME\_CLS, indicating the effectiveness of directly measuring semantic alignment between documents and question dimensions.

\paragraph{Effect of Ontology Grounding.}
Removing ontology grounding causes performance drops on the majority of tasks for both QIME and QIME\_CLS. This indicates that grounding question generation in a biomedical ontology produces more informative and domain-specific dimensions for the evaluated biomedical tasks. 

The final question bank also exhibits broad coverage of UMLS concepts. Each of the 8,855 retained questions is linked to at least one UMLS biomedical concept, with 9,410 unique UMLS terms represented overall. Together with the ablation results, this suggests that ontology grounding is reflected in both the content of the resulting dimensions and their downstream performance.

\paragraph{Effect of Diversity-Aware Dimension Selection.}
Comparing QIME with QIME w/o MMR shows that diversity-aware selection improves performance on 10 out of 12 tasks, indicating that MMR reduces redundancy among activated questions and encourages the embedding to capture complementary semantic aspects of the input.

\subsection{Effect of Top-$k$ Dimension Activation}
\label{sec:topk}
Figure~\ref{fig:topk} examines how performance varies with the top-$k$ activation parameter $k \in \{2^i \mid i=3,\dots,10\}$. We compare the full QIME model with QIME w/o MMR, while the classifier-based variant QIME\_CLS is shown for reference.

Across tasks, QIME's performance remains stable across a broad range of $k$, typically peaking around $128$–$256$. All reported results use a fixed $k=256$ to ensure fair comparison across benchmarks. Notably, QIME often matches or surpasses the classifier-based variant using only a few hundred active dimensions, showing that sparse, diversity-aware activation provides an effective balance between performance and interpretability.


\section{Interpretability Analysis}

Beyond embedding quality, we evaluate the interpretability of QIME's representations through cognitive load analysis and a qualitative case study.

\subsection{Cognitive Load}

Following CQG-MBQA~\cite{DBLP:conf/iclr/SunHTTY25}, we measure cognitive load as the expected number of shared activated dimensions between two documents, corresponding to the inner product of their binary embedding vectors. A lower cognitive load indicates fewer dimensions are required to explain a similarity judgment. 


Table~\ref{tab:cognitive_load} reports cognitive load across all retrieval datasets for question-based interpretable methods. QIME achieves substantially lower cognitive load than all baselines, averaging 136 shared dimensions per document pair compared to 583 for CQG-MBQA and 1,908 for QA-Emb, representing reductions of 76.7\% and 92.9\% respectively. This indicates that QIME's sparse representations impose less cognitive burden on users seeking to understand similarity judgments.


\subsection{Case Study: Interpreting Question-Based Representations}
\label{sec:case}

Table~\ref{tab:case_study} compares the top-ranked embedding dimensions produced by different interpretable methods for the same clinical input describing chest pain in a lung cancer patient. For LDIR-500, the highest-scoring dimensions correspond to long anchor texts, including personal anecdotes and non-biomedical content, providing limited direct insight into which factors drive the representation. CQG-MBQA produces question-based dimensions, but the top-ranked questions are generic and fail to capture domain-specific distinctions. In contrast, QIME activates atomic, ontology-grounded questions that are easier to inspect and more closely aligned with the clinical themes of this input, such as CT-based cardiovascular diagnosis and cancer-related pathology. 



\section{Conclusion}
\label{sec:conclusion}

We introduce \textbf{QIME}, an ontology-grounded framework for constructing question-based interpretable biomedical text embeddings. By grounding question generation in UMLS concept signatures, QIME discovers explicit, domain-specific semantic dimensions. Embeddings are then constructed via similarity-based semantic activation with diversity-aware dimension selection through MMR, yielding sparse and coherent representations without classifier training. Experiments across clustering, semantic similarity, and retrieval benchmarks demonstrate that QIME outperforms prior interpretable embedding methods and substantially narrows the performance gap to strong black-box biomedical encoders. Interpretability analysis further shows that QIME reduces cognitive load by over 76\% relative to existing question-based encoders, providing an effective foundation for transparent and auditable biomedical NLP applications.



\section*{Limitations}

QIME relies on external biomedical resources: a corpus for clustering and an ontology for grounding question generation. As with other knowledge-grounded NLP systems, the coverage and currency of these resources may influence the resulting question dimensions. Meanwhile, QIME currently uses ontology concepts as grounding signals without explicitly exploiting the hierarchical structure. Incorporating ontology relations is an important future direction. UMLS covers the broad biomedical domain, and ontology grounding alone does not establish clinical relevance. Our evaluation primarily measures embedding quality on biomedical benchmarks rather than clinical utility, while validating the usefulness of QIME dimensions in clinical workflows requires expert assessment and task-specific studies. 

Second, activated questions in QIME indicate semantic relevance, which is the primary objective of embedding models, rather than factual entailment. They are intended to support embedding inspection and error analysis, while downstream clinical decision support would require additional task-specific validation.

Finally, there remains a gap between QIME and dense encoders on some retrieval tasks. This reflects a design trade-off: while sparse question-based embeddings may capture less fine-grained signals than dense representations, they provide transparent, semantically explicit biomedical-domain dimensions that support interpretability and auditability.


\section*{Acknowledgments}
This research is supported by the National Research Foundation, Singapore under its AI Singapore Programme (AISG Award No: AISG3-RP-2022-029).

\bibliography{custom}

\newpage
\appendix

\section{Prompt Templates}
\label{sec:appendix_prompt}

We use large language models to generate contrastive, ontology-grounded yes/no questions for each semantic cluster. The prompt below illustrates the template used for ontology-grounded question discovery from semantic clusters.

\begin{tcolorbox}[colback=gray!5, colframe=gray!50!black, title=Prompt: Ontology-Grounded Question Discovery]
You are an expert in biomedical NLP question generation. Generate 10 simple yet insightful yes/no questions that determine properties of an article, where for all questions, the answer will be "yes" for ALL the positive articles and "no" for ALL the negative articles. Questions must focus on biomedical meaning, such as diseases, symptoms, risk factors, treatments, drugs, genes and chemicals. 

\textbf{Constraints:}
\begin{itemize}\setlength{\itemsep}{0pt}\setlength{\topsep}{0pt}
\item Avoid \textbf{meta-level} questions (e.g., "Does the article report results from a clinical trial?", "Does the article discuss methods?"). 
\item IMPORTANT: Output ONLY the numbered questions, no analysis or explanation.
\item Format the questions in a numbered list as shown below: 
1. [First simple yes/no question] 
2. [Second simple yes/no question] 
\end{itemize}

\textbf{Positive Articles:}

Positive 1. \{positive\_chunk\_1\} 

Positive 2. \{positive\_chunk\_2\} 

... 

Positive N. \{positive\_chunk\_N\} \\

\textbf{Negative Articles:}

Negative 1. \{negative\_chunk\_1\} 

Negative 2. \{negative\_chunk\_2\} 

... 

Negative M. \{negative\_chunk\_M\} \\

\textbf{UMLS Context:} 

\{CUI 1: description\_1\} 

\{CUI 2: description\_2\}

...

\end{tcolorbox}

\begin{table*}[t]
\centering
\small
\begin{tabular}{llccccccc}
\toprule
\multirow{2}{*}{\textbf{Model}} & 
\multirow{2}{*}{\textbf{Backbone}} & 
\multicolumn{6}{c}{\textbf{Clustering ( V-Measure $\uparrow$ )}} & \textbf{STS ( SC $\uparrow$ )} \\
\cmidrule(lr){3-8} \cmidrule(lr){9-9}
& &
\textbf{BioP2P} &
\textbf{BioS2S} &
\textbf{MedP2P} &
\textbf{MedS2S} &
\textbf{ClusTREC} &
\textbf{Average} &
\textbf{BIOSSES} \\

\midrule
\textbf{QIME} & Qwen3-30B & 40.37 & 36.78 & 33.92 & 31.44 & 81.99 & 44.90 & 79.66 \\
\textbf{QIME\_CLS} & Qwen3-30B & 38.18 & 34.82 & 33.61 & 32.00 & 79.43 & 43.61 & 61.88 \\
\midrule
\textbf{QIME} & LLaMA-3-70B & 40.24 & 35.92 & 33.90 & 31.59 & 80.86 & 44.50 & 78.48 \\
\textbf{QIME\_CLS} & LLaMA-3-70B & 37.65 & 34.29 & 33.18 & 31.37 & 80.53 & 43.41 & 60.55 \\
\midrule
\textbf{QIME} & GPT-5-mini & 40.09 & 36.67 & 34.08 & 31.79 & 80.96 & 44.72 & 75.05 \\
\textbf{QIME\_CLS} & GPT-5-mini & 36.75 & 34.07 & 33.41 & 31.39 & 80.69 & 43.26 & 59.65 \\
\bottomrule
\end{tabular}
\caption{Clustering and STS performance across different LLM backbones used for question generation. Performance remains consistent across backbones.}
\label{tab:backbone_cluster}
\end{table*}

\begin{table*}[t]
\centering
\small
\begin{tabular}{llccccccc}
\toprule
\multirow{2}{*}{\textbf{Model}} &
\multirow{2}{*}{\textbf{Backbone}} &
\multicolumn{7}{c}{\textbf{Retrieval ( nDCG@10 $\uparrow$ )}} \\
\cmidrule(lr){3-9}
& &
\textbf{NFCorpus} &
\textbf{PHQA} &
\textbf{MedQA} &
\textbf{COVID} &
\textbf{R2-IYI} &
\textbf{R2-PMC} &
\textbf{Average} \\
\midrule
\textbf{QIME} & Qwen3-30B & 25.09 & 75.64 & 62.36 & 64.65 & 11.79 & 7.08 & 41.10 \\
\textbf{QIME\_CLS} & Qwen3-30B & 15.74 & 61.74 & 54.66 & 46.31 & 8.94 & 5.53 & 32.15 \\
\midrule
\textbf{QIME} & LLaMA-3-70B & 24.85 & 75.30 & 62.91 & 63.47 & 11.54 & 8.12 & 41.03 \\
\textbf{QIME\_CLS} & LLaMA-3-70B & 16.15 & 64.95 & 55.51 & 47.57 & 8.78 & 5.14 & 33.02 \\
\midrule
\textbf{QIME} & GPT-5-mini & 25.32 & 73.42 & 61.47 & 60.92 & 11.16 & 6.73 & 39.84 \\
\textbf{QIME\_CLS} & GPT-5-mini & 16.48 & 66.20 & 57.04 & 47.73 & 9.27 & 5.27 & 33.66 \\
\bottomrule
\end{tabular}
\caption{Retrieval performance across LLM backbones used for question generation. Results remain stable across backbones.}
\label{tab:backbone_retrieval}
\end{table*}

\section{Post-Processing of Generated Questions}
\label{sec:appendix_post}

To ensure discriminative and reliable question dimensions, we apply the following post-processing procedure to the generated questions.

\paragraph{Sampling Strategy.}
For post-processing, we use a separate answer-probing sample to estimate whether each generated question discriminates the target cluster from negative examples. For each cluster, we sample $p_{\mathrm{p}}=5$ positive documents from the cluster, $p_{\mathrm{hard}}=3$ hard negatives from the nearest clusters, and $p_{\mathrm{easy}}=2$ easy negatives from randomly selected distant clusters. 

\paragraph{Answer Probing.}
An LLM is used to answer each question for all sampled documents. Responses are normalized to binary \texttt{yes}/\texttt{no} labels.

\paragraph{Discrimination Scoring.}
Each question is assigned a discrimination score:
\begin{equation}
\text{score} =
\frac{\text{yes}_{\mathrm{pos}}}{\text{yes}_{\mathrm{pos}}+\text{no}_{\mathrm{pos}}}
-
\frac{\text{yes}_{\mathrm{neg}}}{\text{yes}_{\mathrm{neg}}+\text{no}_{\mathrm{neg}}}.
\end{equation}

Questions are ranked by this score in descending order.

\paragraph{Redundancy Filtering.}
To remove near-duplicate questions, we compute cosine similarity between question embeddings and retain only questions with cosine similarity below a threshold $\theta=0.8$. For each cluster, up to $adapt_{\mathrm{t}}$ questions are retained, where $adapt_{\mathrm{t}}$ scales with cluster size: 2, 4, 6, and 8 questions for clusters of size ${<}1\text{k}$, $1\text{k}$--$2\text{k}$, $2\text{k}$--$3\text{k}$, and ${>}3\text{k}$, respectively, since larger clusters tend to encompass a broader range of biomedical semantics.

\paragraph{Global Deduplication.} After per-cluster filtering, a final deduplication pass is applied across all clusters: any question with cosine similarity above $\theta{=}0.8$ to any previously retained question is removed. This yields the final global question set $\mathcal{Q} = \{q_1, \ldots, q_M\}$ used for embedding construction.

\section{Classifier-Based Embedding Construction (QIME\_CLS)}
\label{sec:appendix_cls}

For comparison with the proposed similarity-based semantic activation, we implement a classifier-based embedding construction strategy, denoted as \textbf{QIME\_CLS}. In this formulation, each question dimension is modeled as a binary prediction task.

Given a document $x$ and a question $q_j$, the embedding value for dimension $j$ is predicted by a dedicated binary classifier that determines whether the semantic property expressed by $q_j$ is present in $x$. This approach replaces semantic activation of \textbf{QIME} with supervised classification.

In the classifier-based approach, we associate each embedding dimension with a binary classifier corresponding to a question. For each question, we construct 1{,}000 training instances by sampling 300 positive examples from the corresponding cluster, 500 hard negatives from the nearest clusters, and 200 random negatives.

We attach one classification head per question on top of a shared backbone encoder, freeze the backbone parameters, and train only the classification heads. Training is formulated as a multi-task classification problem: given a document--question pair, only the corresponding head is updated using the cross-entropy loss. We train for 3 million steps with a batch size of 1. After training, the outputs of all classification heads for a given input text are concatenated to form the final classifier-based embedding.

\section{Backbone Sensitivity Analysis}
\label{sec:appendix_backbone}

To evaluate the sensitivity of QIME to the choice of LLM backbone used for question generation, we additionally generate question sets using LLaMA-3-70B and GPT-5-mini, while the main paper reports results using Qwen3-30B. 

Tables~\ref{tab:backbone_cluster} and \ref{tab:backbone_retrieval} report the results across clustering, biomedical STS and retrieval benchmarks. Across all tasks, performance remains highly consistent across different LLM backbones. In particular, QIME maintains similar improvements over the classifier-based variant regardless of the backbone used.

These results suggest that the performance gains of QIME primarily arise from ontology-grounded question construction and similarity-based semantic activation with diversity-aware dimension selection, rather than reliance on a specific LLM backbone.

\end{document}